\documentclass[runningheads]{llncs}
\usepackage{color}
\usepackage{colortbl}
\usepackage{hyperref}
\usepackage{graphicx}
\usepackage{float}
\definecolor{LightGray}{rgb}{0.9,0.9,0.9}
\usepackage{graphicx}
\begin{document}
\title{CYRUS Soccer Simulation 2D Team Description Paper 2021}
%
%
\author{Nader Zare\inst{1}\and Aref Sayareh\inst{3}\and Mahtab Sarvmaili\inst{1}\and Omid Amini\inst{4}\and Am\'ilcar Soares\inst{1}\and Stan Matwin\inst{1}\inst{2}}
\authorrunning{N. Zare et al.}
%
\institute{
Institute for Big Data Analytics, Dalhousie University, Halifax\\
\and
Institute for Computer Science, Polish Academy of Sciences, Warsaw\\
\and
Shiraz University, Iran\\
\and
Qom University of Technology, Iran\\
\email{\{nader.zare, mahtab.sarvmaili, amilcar.soares\}@dal.ca}\\ 
\email{stan@cs.dal.ca}\\
\email{\{arefsayareh, omidamini\}@gmail.com}\\
}
\maketitle              

\begin{abstract}
In this report, we briefly present the technical procedure and simulation steps for the 2D soccer simulation of team Cyrus. 
We emphasize on this document on how the prediction of teammates' behavior is performed. 
In our proposed method, the agent receives the noisy inputs from the server, and predicts the ball holder full state behavior. 
Taking advantage of this approach for choosing the optimal view angle shows 11.30\% improvement on the expected win rate.
\end{abstract}
\sloppy
\section{Introduction} 
The idea of robotic soccer games was proposed as a novel research topic in 1992, and since then the RoboCup has been considered as an annual international competition for developing new ideas in A.I. and robotics.  
This competition is formed of various leagues such as Rescue\cite{resque1,resque2,resque3,resque4,resque5}, Soccer Simulation\cite{3d} and Standard Platform\cite{stand} leagues.

Cyrus Team is one of the soccer simulation team in the 2D Soccer Simulation league. 
This team was established in 2013, and it has engaged in RoboCup and IranOpen competitions since then. 
It is worth mentioning that this team has gained the second, third, fourth, and fifth places in RoboCup 2018, 2019, 2017, 2014 years respectively. 
Also, Cyrus won first place in IranOpen 2018 and 2014, RoboCup Asia-Pacific 2018, and second place in JapanOpen 2020 competitions. 
The Cyrus’s team base is agent2d\cite{agent2d}.

\subsection{Previous Work}
In the recent years we have concentrated on exploiting artificial intelligence and machine learning techniques to improve the functionality of Cyrus team \cite{cyrus14,cyrus15,cyrus18,cyrus19}. 
Among these works, we can mention the improvement of agents' defense decision-making process using Reinforcement Learning (RL)\cite{rl1}, prediction of an opponent's behavior, and optimization of the shoot skill. 
Helios has developed an algorithm for the analysis of the agents' offensive behavior \cite{helios19,helios18,heliospaper}. 
Fractals, 2019 which is partially based on Gliders2d used elements of evolutionary  computation, within the framework of Guided Self-Organisation \cite{fractals}. 
FRA-United has researched on the commutation of agents in games \cite{fra16,fra18,fra19}. FCP\_GPR teams has developed a framework for the free-kick \cite{fcp}, while the Namira has implemented a python-based application for the analysis of soccer simulation games\cite{namira1,namira2,namira3}. 
Razi has worked on scoring the offensive behavior in the 2D soccer simulation\cite{razi18,razi19}.            
\setcounter{footnote}{0} 

\subsection{Release}
\subsubsection{Cyrus 2014 Source}

As a part of our contribution to the development of the 2D Soccer Simulation league, we have released the Cyrus 2014 \cite{cyrus14} source code to encourage new teams to participate in the competitions. 
Cyrus 2014 won the 1st and 5th places in the Iran-Open RoboCup Competition and International RoboCup Competition, respectively. 
The source code can be found in our github\footnote{Cyrus 2014 Source \url{https://github.com/naderzare/cyrus2014}.}.

\subsubsection{Starter Agent and Starter Librcsc}
Cyrus team members - in cooperation with IranOpen technical committee of 2D soccer simulation league - have designed a simplified version of the agent base \cite{agent2d} and the librcsc library for the 2D soccer simulation starter league. 
High-level behaviors like passing, dribbling, and shooting have been omitted from this base. 
This version of 2D soccer simulation base and librcsc - specifically are designed for junior students - have been exploited in 2D soccer simulation starter league during both IranOpen RoboCup 2018, IranOpen RoboCup 2020 and RoboCup Asia-Pacific 2018. 
More than ten teams participated in each of the competitions, with more than fifty participants in total.
All of the participants have used the this base developed by Cyrus and IranOpen committee of 2D soccer simulation league. 
The base can be found in our github\footnote{Starter Agent 2D \url{https://github.com/naderzare/StarterAgent2D}} \footnote{Starter LibRCSC \url{https://github.com/naderzare/StarterLibRCSC}}.

\subsubsection{CppDNN}
The C++ Deep Neural Network (CppDNN) library has been developed by Cyrus team members to facilitate the implementation of Deep Neural Network in the 2D Soccer Simulation environment. 
This library stores the trained weights of a neural network which has been trained by Keras library. 
The developed script within this library transforms the trained weights of a deep neural network into a text file. 
Subsequently, it loads the trained weights to recreate the original deep neural network in C++. 
This library employs Eigen Library for its calculation. 
The library can be found in our github\footnote{CppDNN Source Code \url{https://github.com/Cyrus2D/CppDNN}}.
\subsubsection{Pyrus - Python Soccer Simulation 2D Base}
Most of 2D soccer simulation teams exploit the Helios \cite{agent2d}, Gliders2d \cite{gldbase}, WrightEagle \cite{wrbase} or Oxsy \cite{oxsy} base. 
All of these bases have been developed in C++. 
Although those have shown fast processing and execution time, developing machine learning algorithms will be a challenging and time-consuming process.  
Due to the fast growth of Python language popularity among students and scientist, and its strength for implementing machine learning algorithms, Cyrus team members have started developing an open source python base for 2D soccer simulation league. 
This base is currently available in Cyrus github\footnote{Pyrus Base Source Code \url{https://github.com/Cyrus2D/Pyrus}} and it will support all features of current 2D soccer simulation server in the Full-State mode in the near future.

\section{Kick Behavior Predictor}
\begin{figure}[h!]
    \includegraphics[scale=0.35]{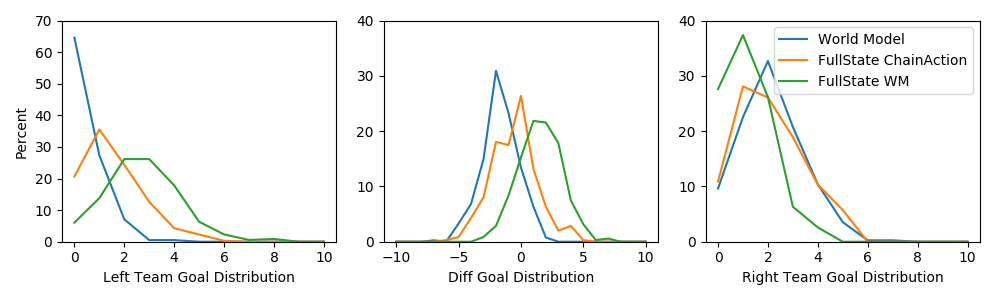}
    \caption{The distribution of goal for and goal against between Cyrus and Helios in three different modes: a. WM b. FWM and c. fullstate chain action .}
    \label{fig:Fig1}
\end{figure}

One of the main goals of 2D Soccer team is increasing the winning chance, and it can be achieved by enhancing the general performance of the team. 
This objective can be interpreted as increasing the team's number of goals and reducing the number of goals against the team. 
Enhancing the functionality of the team's results in a better performance in the field.
However, random noises on the observation of the agents from the environment are the major challenge the agents face while they want to choose their actions. 
The 2D soccer environments exert the random noises on the observation of agents from the environment to simulate the real-world soccer match; however, these noises complicate the agents' decision-making process. 
The soccer simulation server provides an option known as \textbf{\emph{"full-state mode"}} to eliminate the random noises from the the agents' observation. 
If the server runs with the \textbf{\emph{"full-state mode"}}, it distributes the pure state of the game to the teams. 
In order to understand the impact of noise on the functionality of teams, we tested the Cyrus against Helios 2019\cite{helios19} with two different settings for the simulation server. 
In the first, server was run with its default settings. 
In the second, the server was run with the full-state mode. 
This phase was divided into two sub-experiments. 
Cyrus receives the full-state of the game from the server and uses it in two different fashions: 1 - full-state observation: the agents exert the  pure observation of the system for their decision-making; 2 - full-state chain action: the pure observations are only exploited for the chain action of agents, and the noisy world model was used for the rest of processing. 
These three operation modes have been tested 500 times, and the experimental results are reported in the following section. The distribution of goal for and goal against for these experiments are shown in Fig. \ref{fig:Fig1}.
Also, the win rate, expected win rate, and average score are denoted in Table\ref{table:Table1}.
The results of these experiments prove the extreme effect of noisy data on the functionality of the teams. 
In order to tackle this problem, many team are exploiting opponent behavior prediction or noise reduction algorithms.
In this TDP, we aim to address this challenge by enabling agents to predicts their full-state behavior using the noisy observations and exploiting this prediction for the optimization of their behavior.        
Correspondingly, the server ran in the fullstate mode, and it passed the word Model (WM) and the Fullstate World Model (FWM) to the agents. 
At this point the WM and FWM will be received by the agent for the further processing. 
The agent passes the FWM and WM to the Kick Decision-Making module, and it only passes the WM to Move Decision-Making module. 
If the ball is not within the kickable area of the agent, move-decision module chooses behavior of agent and it sends the action to the server, otherwise kick-decision making module sends the WM and FWM to Data Extractor module and Chain action module respectively. 
Rhe Chain Action module employs the FWM to choose optimal action, then afterward, it sends the action and to Data Extractor module and server.   

\begin{figure}[h!]
    \includegraphics[scale=0.23]{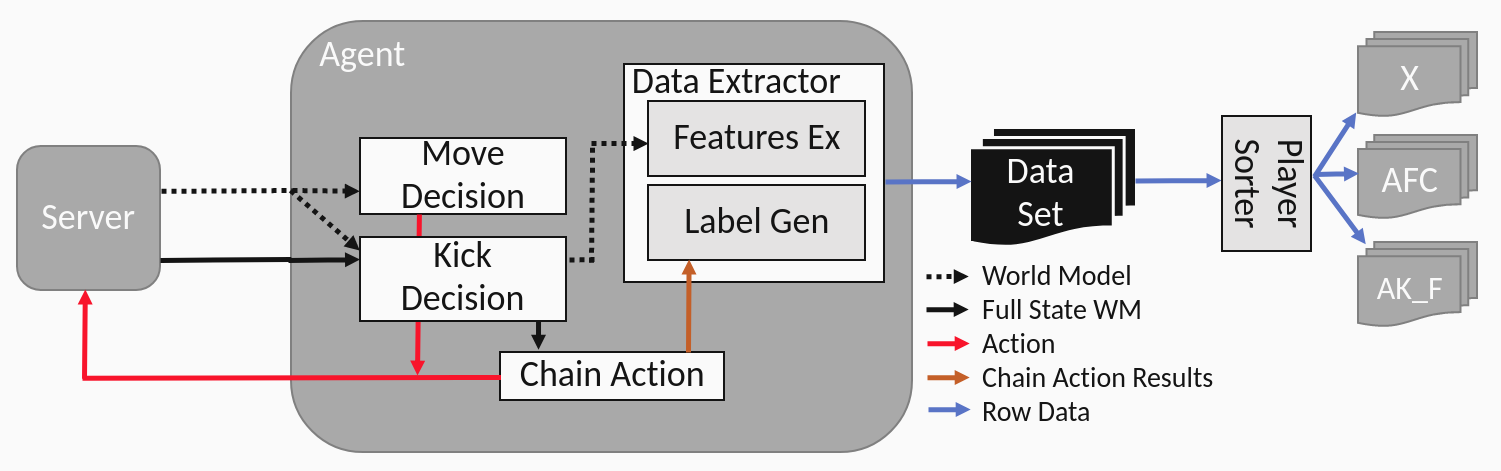}
    \caption{The internal structure of agent and its processing modules}
    \label{fig:Fig2}
\end{figure}

The Data Extractor Module receives the WM and action and it attempts to generate the Data Set using its submodules (Feature Extractor and Label Generator). 
Feature extractor is a part of the data extractor module to select the important features (will be discussed in the next subsection) from the received data. 
Label Generator takes the action of the agent from Chain Action module and generates the fullstate action label for this data. 
The structure of agent and its processing modules are presented in the Fig. \ref{fig:Fig2}.

\begin{table}[]
    \tiny
    \centering
    \caption{The win rate, expected win rate, and average score}
    \label{table:Table1}
    \begin{tabular}{|l|l|l|l|l|}
        \hline
        Run Type & Win Rate & Expected Win Rate & Cyrus Average Goal & Helios2019 Average Goal \\
        \hline
        Normal & 7.09 & 8.19 & 0.45 & 2.13 \\
        Chain Action Full State & 24.64 & 33.46 & 1.59 & 2.09 \\
        Full State & 72.7 & 85.76 & 2.72 & 1.19 \\
        \hline
    \end{tabular}
\end{table}

\subsection{Feature Extractor}

As we mentioned in Section 2, the feature extractor module receives the WM, and it extracts the most significant attributes of the input data. 
The related features of the ball, players, and others are denoted in Tables \ref{table:Table2} \ref{table:Table3} respectively.

\begin{table}[h]
    \tiny 
    \centering
    \caption{List of Ball Features and Other}\label{tab1}
    \label{table:Table2}
    \begin{tabular}{|l|l|l|}
        \hline
        Feature Class & Feature Name &  Description \\
        \hline
        Ball\_Position & Ball\_X &  Ball Position - X \\
        Ball\_Position & Ball\_Y &  Ball Position - Y \\
        Ball\_Position & Ball\_RX &  Distance to Holder Player - X \\
        Ball\_Position & Ball\_RY &  Distance to Holder Player - Y \\
        Ball\_Position & Ball\_R &  Euclidean Distance from Ball to Holder Player \\
        Ball\_Position & Ball\_Teta & Angle From Holder Player to Ball \\
        Ball\_Velocity & Ball\_VX & Ball Velocity - X \\
        Ball\_Velocity & Ball\_VY & Ball Velocity - Y \\
        Ball\_Velocity & Ball\_VR & Ball Velocity - Length \\
        Ball\_Velocity & Ball\_VTeta & Ball Velocity - Angle \\
        Dribble & Dribble\_Free\_Distance & Distance of ball to the nearest opponent in 12 sector \\
        Other & Cycle & Cycles of the game \\
        Other & Offside count & The accuracy count for the offside line\\
        \hline
    \end{tabular}
\end{table}
\begin{table}
    \tiny 
    \centering
    \caption{List of player's features}\label{tab1}
    \label{table:Table3}
    \begin{tabular}{|l|l|l|l|}
        \hline
        Feature Class & Feature Name & Tm or Opp &  Description \\
        \hline
        Other & Player\_Side & Both & Side of player 1 or -1 \\
        Other & Player\_Unum & Both & Uniform number of player \\
        Other & Player\_Body & Both & Body angle \\
        Other & Player\_Face & Both & Face angle \\
        Other & Player\_Tackling & Both & Is player tackling \\
        Other & Player\_Kicking & Both & Is player kicking \\
        Other & Player\_Card & Both &  Has player yellow card or no \\
        Type & Player\_Type\_DashRate & Both &  Dash Rate of player\\
        Type & Player\_Type\_EffortMax & Both & Maximum Effort of player \\
        Type & Player\_Type\_EffortMin & Both & Minimum Effort of player \\
        Type & Player\_Type\_KickableDist & Both & Kickable Distance of player \\
        Type & Player\_Type\_MarginDist & Both & Margin Distance of player \\
        Type & Player\_Type\_KickPowerRate & Both & Kich Power rate of player\\
        Type & Player\_Type\_Decay & Both & Decay of player\\
        Type & Player\_Type\_Size & Both & Size of player \\
        Type & Player\_Type\_SpeedMax & Both & Maximum speed of player \\
        Position & Player\_X & Both & Position of player - X \\
        Position & Player\_Y & Both & Position of player - Y\\
        Position & Player\_RX & Both & Distance to holder player - X\\
        Position & Player\_RY & Both & Distance to holder player - Y\\
        Position & Player\_R & Both & Distance of player to holder player \\
        Position & Player\_Teta & Both & Angle from holder player to player \\
        Position & Player\_Offside & Teammate & Player is in offside \\
        Velocity & Player\_VX & Both & Velocity of player - X\\
        Velocity & Player\_VY & Both & Velocity of player - Y \\
        Velocity & Player\_VR & Both & Velocity of player - Length \\
        Velocity & Player\_VTeta & Both & Velocity of player - angle\\
        Position & Player\_PosCount & Both & Count since last position observation \\
        Velocity & Player\_VelCount & Both & count since last velocity observation \\
        Other & Player\_IsKicker & Teammate & Is this player kicker \\
        Pass & Player\_FreePassAngle & Teammate & Maximum free angle for direct pass \\
        Pass & Player\_DirectPassDist & Teammate & Distance from ball to player \\
        Opponent & Player\_NearestOpponentDist & Teammate & Minimum distance from opponent to player \\
        Position & Player\_GCA & Both & Angle from player to opponent goal center \\
        Position & Player\_GCD &  Both & Distance from player to opponent goal center \\
        Shoot & Player\_FreeShootAngle & Teammate & Maximum free angle for shoot \\
        Stamina & Player\_Stamina & Both & Stamina of player \\
        Stamina & Player\_StaminaCount & Both & Count since last stamina observation \\
        \hline
    \end{tabular}
\end{table}
\subsection{Features Sorting Methods}
\begin{figure}[h!]
    \centering
    \includegraphics[scale=0.15]{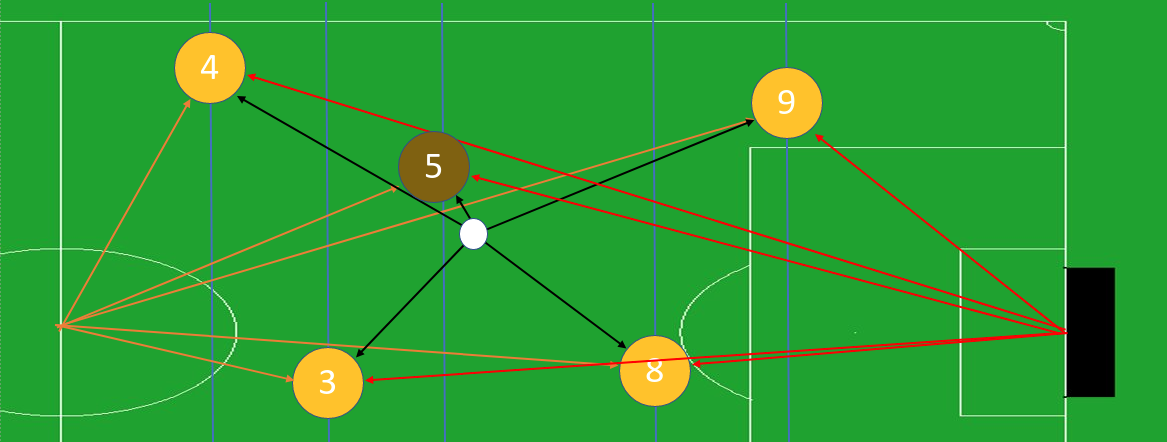}
    \caption{Sample positions of agents in the field}
    \label{fig:Fig3}
\end{figure}

In our proposed method, we take advantage of a deep neural network for the prediction of the agents' behavior using the noisy observations. We've generated 10 different datasets from the world model to examine the effect of the input setting on the prediction of the network. To create each one of this dataset, we used one of the sorting method that is explained in Table  \ref{table:Table4}. Each one of this sorting method changes the order of players' features. To make the process of these sorting methods more clear, the results of them for the players in Fig. \ref{fig:Fig3} are demonstrated in Table \ref{table:Table4}.

\begin{table}[]
    \tiny 
    \centering
    \caption{Sorting Algorithm}\label{tab1}
    \label{table:Table4}
    \begin{tabular}{|l|l|}
    \hline
    Sorting Method & Description \\
    \hline
        \rowcolor{LightGray}
         X & Sorting players of each team by their X of position \\
         \rowcolor{LightGray}
         & Sorting Results: 9 8 5 3 4 \\
         X\_FK & Similar to X approach, but the Kicker player has the first place in sorting\\
         & Sorting Results: 5 9 8 3 4 \\
         \rowcolor{LightGray}
         Unum & Sorting players of each team by their Uniform Number \\
         \rowcolor{LightGray}
         & Sorting Results: 3 4 5 8 9 \\
         Unum\_FK & Like X, But Kicker Player be first \\
         & Sorting Results: 5 3 4 8 9 \\
         \rowcolor{LightGray}
         AFC & Sorting player of each team by their angle from their current position to center of field \\
         \rowcolor{LightGray}
         & Sorting Results: 4 5 9 8 3 \\
         AFC\_FK & Similar to AFC, but the Kicker player has the first place in sorting\\
         & Sorting Results: 5 4 9 8 3 \\
         \rowcolor{LightGray}
         AK & Sorting player of each team by their angle from their current position to the kicker player \\
         \rowcolor{LightGray}
         & Sorting Results: 4 5 9 8 3 \\
         AK\_FK & Similar to AK, but the Kicker Player has the first place in sorting \\
         & Sorting Results: 5 4 9 8 3 \\
         \rowcolor{LightGray}
         AKG & Sorting Player of each Team by angle from position to Goal Center \\
         \rowcolor{LightGray}
         & Sorting Results: 9 4 5 3 8 \\
         AKG\_FK & Similar to AK, but the Kicker player has the first place in sorting\\
         & Sorting Results: 5 9 4 3 8 \\
    \hline
    \end{tabular}
\end{table}
\subsection{Label Generator}
This module takes action from the chain action module to generate the labels for each the data raw. The labels of data are noted in Table \ref{table:Table5}.
\begin{table}
    \tiny
    \centering
    \caption{List of Labels}\label{tab1}
    \label{table:Table5}
    \begin{tabular}{|l|l|l|}
        \hline
        Label &  Description \\
        \hline
        Category &  Hold \(\vert \vert\) Pass \(\vert \vert\) Dribble \\
        TargetUnum & Uniform number of target player \\
        TargetIndex & Index of target player after sorting\\
        Description &  Dribble \(\vert \vert\) Direct Pass \(\vert \vert\) Cross Pass \(\vert \vert\) Through Pass \(\vert \vert\) Lead Pass \\
        TargetPosition &  Target position \\        
        FirstKickAngle & Angle of selected action from ball \\
        FirstKickSpeed & Ball kick speed \\
        \hline
    \end{tabular}
\end{table}
\subsection{Results}

\begin{figure}[H]
    \centering
    \includegraphics[scale=0.25, trim= 1.2cm 15cm 0.8cm 14cm, clip]{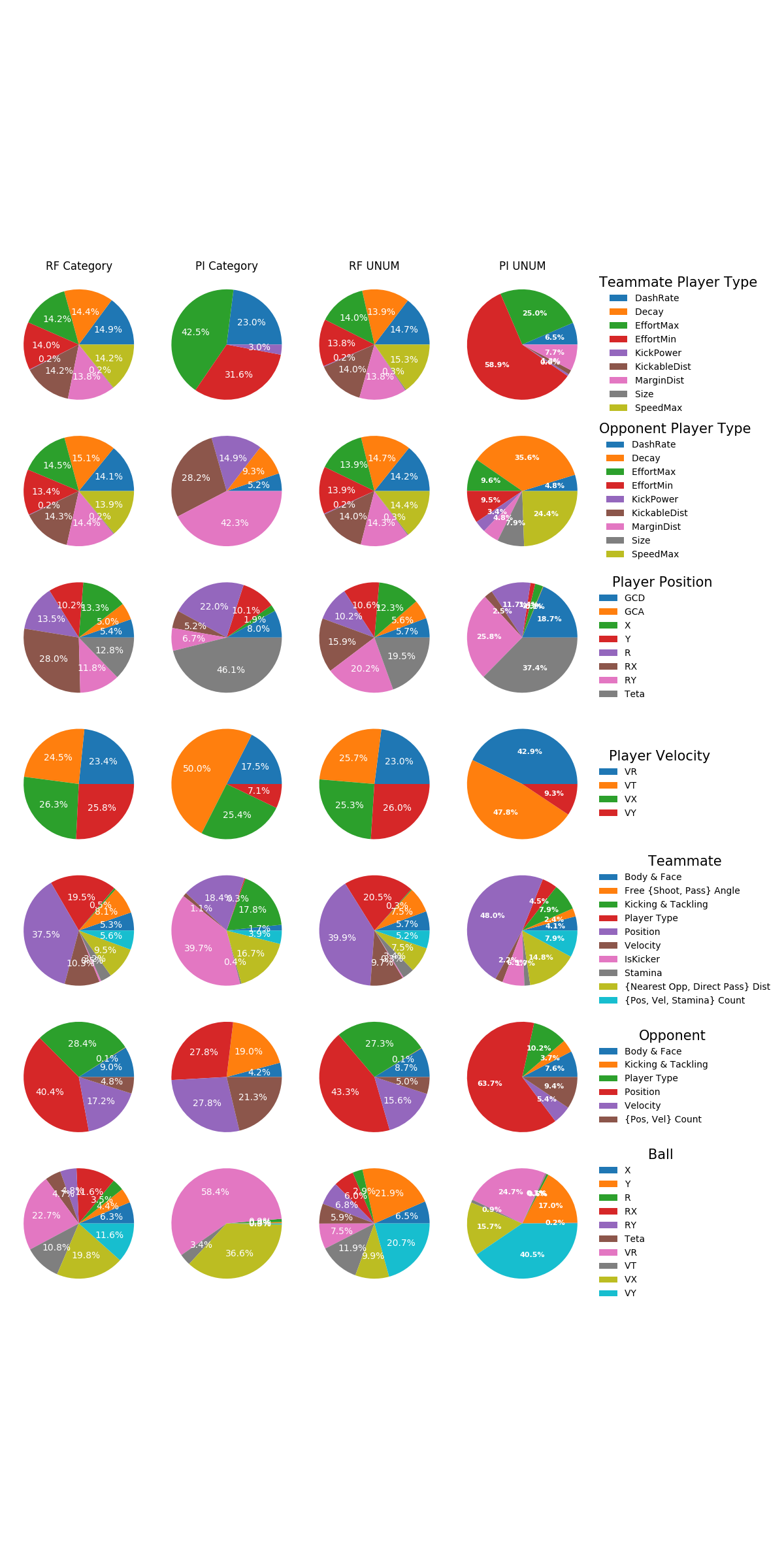}
    \caption{The important features of sorted data extracted by Random Forest and Permutation Feature Importance}
    \label{fig:Fig4}
\end{figure}

\begin{figure}[h!]
    \includegraphics[scale=0.276]{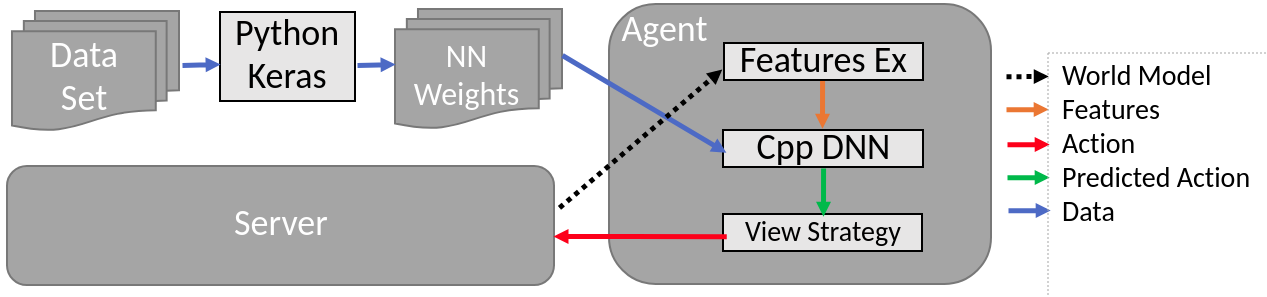}
    \caption{The process of behavior prediction}
    \label{fig:Fig5}
\end{figure}

To examine the impact of different input features on the prediction of the neural network, we chose 1 million of the Cyrus and Helios2019 raw data for training the deep neural network. We have created 10 diverse dataset using different sorting methods(see Table \ref{table:Table4}). The whole process of behavior prediction is demonstrated in Fig. \ref{fig:Fig5}. We reported the accuracy and error rate of model for those 10 dataset in Table \ref{table:Table6}. According to Table \ref{table:Table6}, Unum\_FK has better accuracy in comparison to the approaches.  
\begin{table}
    \small
    \centering
    \caption{Accuracy and error rate of the model for 10 datasets}\label{tab1}
    \label{table:Table6}
    \begin{tabular}{|l|l|l|l|l|l|l|l|l|l|l|l|}
        \hline
        Prediction & Type & \rotatebox{75}{X} & \rotatebox{75}{X\_FK} & \rotatebox{75}{Unum} & \rotatebox{75}{Unum\_FX} & \rotatebox{75}{AKG} & \rotatebox{75}{AKG\_FK} & \rotatebox{75}{AK} & \rotatebox{75}{AK\_FK} & \rotatebox{75}{AFC} & \rotatebox{75}{AFC\_FK} \\
        \hline
        \rowcolor{LightGray}
        Category & Classification & 76.55 & 77.23 & 76.69 & \cellcolor{green} 77.37 & 76.09 & 76.61 & 76.08 & 76.63 & 76.03 & 76.41 \\
        Unum & Classification & 57.22 & 57.71 & \cellcolor{green}60.57 & 60.39 & 56.17 & 56.48 & 56.22 & 56.93 & 56.7 & 57.31\\
        \rowcolor{LightGray}
        Unum in Passes & Classification & 57.87 & 58.04 & 61.80 & \cellcolor{green}62.51 & 57.27 & 57.71 & 57.07 & 57.49 & 57.70 & 57.20\\
        Index & Classification & 58.89 & 60.20 & 60.22 & \cellcolor{green}60.84 & 57.79 & 58.45 & 57.08 & 58.66 & 56.51 & 58.57\\
        \rowcolor{LightGray}
        Index in Passes & Classification & 58.49 & 59.73 & 61.79 & \cellcolor{green}62.01 & 58.58 & 58.13 & 58.72 & 58.42 & 57.39 & 57.06\\
        Description & Classification & 71.62 & \cellcolor{green}72.12 & 71.31 & 71.53 & 71.36 & 71.34 & 71.32 & 71.58 & 71.70 & 71.49\\
        \rowcolor{LightGray}
        TargetPosition & Regresion & 2.44 & 2.43 & 2.59 & 2.38 & 2.50 & 2.42 & 2.79 & 2.41 & \cellcolor{green}2.29 & 2.59\\ 
        FirstKickAngle & Regresion & 5.14 & 6.51 & 5.22 & 6.58 & 6.65 & 7.30 & \cellcolor{green}5.11 & 5.36 & 7.34 & 5.14\\
        \rowcolor{LightGray}
        FirstKickSpeed & Regresion & \cellcolor{green}0.041 & 0.054 & 0.043 & 0.054 & 0.055 & 0.06 & 0.042 & 0.044 & 0.061 & 0.042\\
        \hline
    \end{tabular}
\end{table}
In this section we try to evaluate the value of features for the prediction of agents' behavior. To accomplish this task, we exerted the Random Forest algorithm implemented in Sckiti-Learn and Permutation Feature Importance implemented by ELI5 library. The Permutation Feature Importance algorithm attempts the most effective features of input data for a trained neural network. 
We evaluated the value of sorted data features (sorted by UNUM) using Random Forest algorithm . Also, we selected two of our predictor deep neural networks that were trained by the UNUM sorted dataset to predict the Category or UNUM of target teammate. Using these two neural networks and the Permutation Feature Importance algorithm we chose the significant features of data. see Fig.\ref{fig:Fig4}.   
\vspace*{-5pt}
\subsection{How To Use Predictor}
To assess the effect of the fullstate predictor network, we have examined it on Cyrus 2020. The trained neural network - that has been trained by sorted data (UNUM-FK sorted) - predicts the UNUM of fullstate target player. If the ball holder wouldn't have accurate information about the ball receiver agent, it prioritizes observing that agent. The experimental results of this approach on Cyrus team suggest win rate improvement from 8.19\% to 17.49\% and goal rate from 0.45\% to 0.89\%.
\section{Conclusion}
This paper describes all of the previous efforts and current research of Cyrus2020 on the exploitation of AI algorithms in 2D soccer simulation. Using the \textbf{\emph{"fullstate mode"}} of the server, we created a dataset from agents' perceived observations and their FWM behavior. Then we sorted this dataset, and we fed them to the disparate deep neural network for the behavior prediction. Subsequently, we exerted the best trained neural network to optimize the viewpoint of players. The experimental results demonstrate the improvement of the win rate and goal rate.          
\section{Future Work}
In the near future we plan to improve the proposed approach in this TDP. We are planning to exert the Convolution Neural Network (CNN) as our predictor network. For the next step, we intend to process our data using the recurrent neural network which can process temporal data.

\end{document}